\documentclass{article}
\usepackage{spconf,amsmath,graphicx}
\usepackage{enumitem}
\usepackage{multirow}
\usepackage[table,dvipsnames]{xcolor}
\newcolumntype{L}{@{}>{\kern\tabcolsep}l<{\kern\tabcolsep}}
\usepackage{booktabs}
\usepackage{amsfonts}
\usepackage{hyperref}
\usepackage{textpos}

\usepackage[pages=some,placement=top]{background}

\DeclareMathOperator*{\maximum}{\textit{max}}
\DeclareMathOperator*{\minimum}{\textit{min}}

\title{Deep Multi-Modal Classification of Intraductal Papillary Mucinous Neoplasms (IPMN) with Canonical Correlation Analysis}
\name{Sarfaraz Hussein$^1$, Pujan Kandel$^2$, Juan E. Corral$^2$, Candice W. Bolan$^2$,}
\secondlinename{Michael B. Wallace$^2$ and Ulas Bagci$^1$}
\address{$^1$Center for Research in Computer Vision (CRCV), University of Central Florida, Orlando, FL. \\ 
$^2$Mayo Clinic, Jacksonville, FL.}
\begin{document}
%
\backgroundsetup{contents=Accepted for publication in IEEE International Symposium on Biomedical Imaging (ISBI) 2018,color=black!100,scale=1.5,opacity=0.7,position={5.5,1.35}}
\BgThispage

\maketitle
\begin{abstract}
Pancreatic cancer has the poorest prognosis among all cancer types. Intraductal Papillary Mucinous Neoplasms (IPMNs) are radiographically identifiable precursors to pancreatic cancer; hence, early detection and precise risk assessment of IPMN are vital. In this work, we propose a Convolutional Neural Network (CNN) based computer aided diagnosis (CAD) system to perform IPMN diagnosis and risk assessment by utilizing multi-modal MRI. In our proposed approach, we use minimum and maximum intensity projections to ease the annotation variations among different slices and type of MRIs. Then, we present a CNN to obtain deep feature representation corresponding to each MRI modality (T1-weighted and T2-weighted). At the final step, we employ canonical correlation analysis (CCA) to perform a fusion operation at the feature level, leading to discriminative canonical correlation features. Extracted features are used for classification. Our results indicate significant improvements over other potential approaches to solve this important problem. The proposed approach doesn't require explicit sample balancing in cases of imbalance between positive and negative examples. To the best of our knowledge, our study is the first to automatically diagnose IPMN using multi-modal MRI.
\end{abstract}
\section{Introduction}
\label{sec:intro}

Cancer is one of the main causes of death in the world with a mortality rate of 171.2 per 100,000 people per year (based on 2008-2012 stats)~\cite{americancancer}. Among all cancers, pancreatic cancer has the poorest prognosis with a 5-year survival rate of just 7\% in the United States~\cite{americancancer}. To address the problem of automatic diagnosis of pancreatic cancer, we propose a new CAD framework for Intraductal Papillary Mucinous Neoplasms (IPMN). IPMN is a mucin-producing neoplasm found in the main and branch pancreatic ducts. They are radiographically identifiable precursors to pancreatic cancer~\cite{shi2012intraductal,sadot2015tumor}. If left untreated, they can progress into invasive cancer. For instance, around one-third of resected IPMNs are found to be associated with invasive carcinoma~\cite{matthaei2011cystic}. In 2012, Tanaka et al.~\cite{tanaka2012international} published the International consensus guidelines for the preoperative management of IPMN using radiographic and clinical criteria. These guidelines can be used in the development of CAD approaches for the separation of IPMNs from normal pancreas. The CAD approaches can yield to identify important imaging bio-markers that may assist radiologists for improved diagnosis, staging, and treatment planning.

In the literature, there are a limited number of studies addressing the automatic diagnosis of IPMN using radiology images. Hanania et al.~\cite{hanania2016quantitative} studied the contribution of numerous low-level imaging features such as texture, intensity, and shape to perform low and high-grade IPMN classification. In the approach by Gazit et al.~\cite{gazit2017quantification} texture and component enhancing features were extracted from the segmented cysts. The process is then followed by a feature selection and classification framework. Both of these approaches~\cite{hanania2016quantitative,gazit2017quantification}, however, are evaluated on CT images and require the segmentation of cysts or pancreas. In contrast to these methods, our approach doesn't require prior segmentation of cysts or pancreas and is evaluated on multi-modal MRI scans rather than CT. In this work, we hypothesize and evaluate the influence of complementary information in T1-weighted and T2-weighted scans that can be utilized to perform improved diagnosis of IPMN.\\

\noindent \textbf{Our Contributions:}
\begin{itemize}[leftmargin=*]
\itemsep0em 
\item To the best of our knowledge, this is the first study to use deep learning for the classification of IPMN.
\item We employ multiple imaging modalities of MRI (T1 and T2) and fuse the feature representation using Canonical Correlation Analysis (CCA) to obtain better discrimination between normal and subjects with IPMN. We also perform further stratification of IPMN in low-grade and high-grade categories.
\item Extensive experimental evaluations are performed on a
dataset comprising 139 subjects, the largest study of IPMN to date.
\end{itemize}
\vspace{-0.2 in}
\begin{figure*}[h]
\hspace{-0.05in}
\includegraphics[width=180mm]{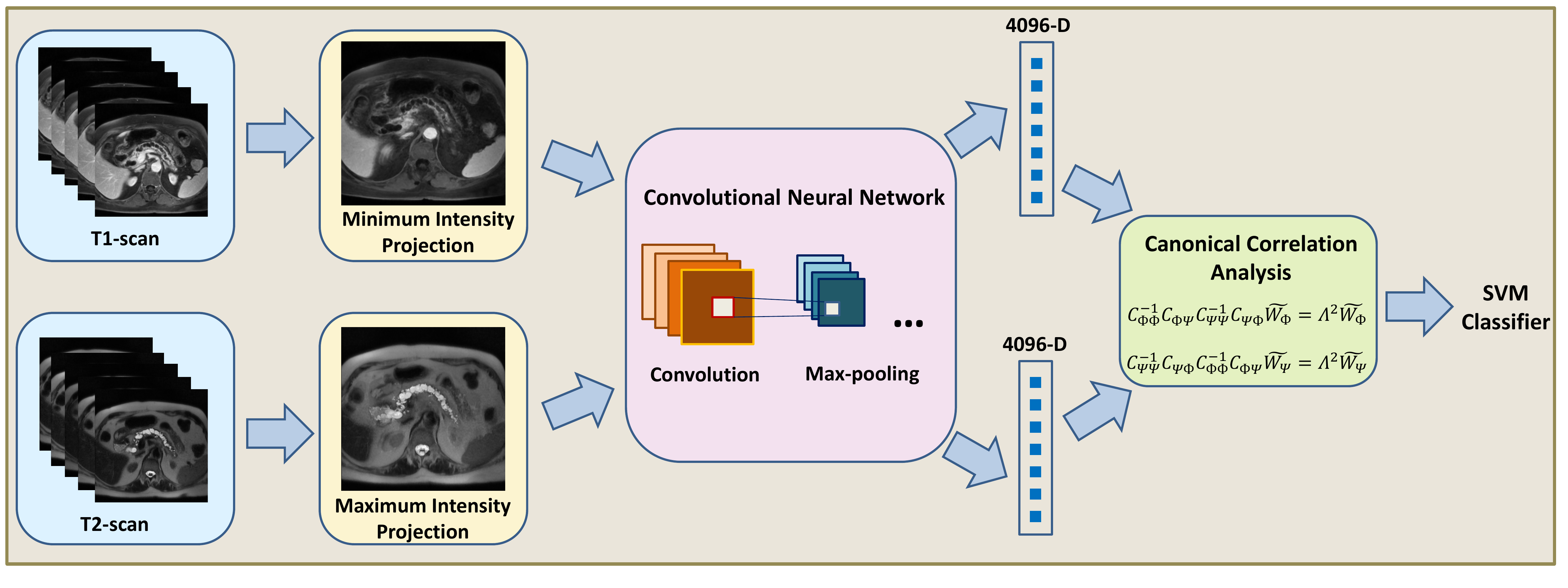}\vspace{-0.4 cm}
\caption{An overview of the proposed method. First, the minimum and maximum intensity projections are computed corresponding to T1 and T2 scans respectively. The intensity projections are then fed into a pre-trained Convolutional Neural Network (CNN) to obtain feature representation. Canonical Correlation Analysis (CCA) based feature fusion is performed in order to obtain discriminative and transformed feature representation. Finally, an SVM based classifier is employed to obtain the final label (normal or IPMN).}
\label{fig:workflow}
\vspace{-0.1 cm}
\end{figure*}

\section{Materials}
\label{sec:mat}
We evaluated our proposed approach for the classification of IPMN on a dataset comprising post-contrast volumetric T1 and T2 MRI scans from 139 subjects. The scans were labeled as normal or IPMN using pathology report obtained after surgery. Out of 139 scans, 108 were from subjects diagnosed with IPMN, whereas the rest of 31 subjects were normal. The in-plane spacing (xy-plane) of T1-weighted scans was ranging from 0.66 mm to 1.48 mm and that of T2-weighted scans from 0.47 mm to 1.41 mm. For pre-processing, we first apply N4 bias field correction~\cite{tustison2010n4itk} to each scan in order to minimize intensity inhomogeneity. Next, we use a curvature anisotropic image filter to smooth images while preserving edge information. For each image, a single slice which has a significant portion of the pancreas is annotated to be normal or IPMN.  

\section{Methods}
\label{sec:method}
\subsection {CNN for Multi-modal Feature Representation:}
\noindent{\textbf{Problem Formulation:}}\\
Our proposed approach consists of inputs from two different MRI image modalities T1 and T2. Let $\mathcal{I}_{T1}=[I_1,I_2 \dots I_{N_1}]\in\mathbb{R}^{X_1 \times Y_1 \times N_1}$ be the T1 scan whereas the corresponding T2 scan is represented as $\mathcal{J}_{T2}=[J_1,J_2 \dots J_{N_2}]\in\mathbb{R}^{X_2 \times Y_2 \times N_2}$, with $N_1$ and $N_2$ number of slices.

Consider $I_u$ be the slice with pancreas from T1 scan and $J_v$ be the slice from T2. Predicting the label from a single slice, however, may yield to hypersensitivity in annotation labels as well as miss important contextual information from the other slices. In order to address these issues, we sample $k=2$ consecutive slices before and after $I_u$ and $J_v$. Since the input to the deep network is a 2D image, we use Maximum and Minimum Intensity Projections to combine information across various slices into a single slice. We employ minimum intensity projection for T1 scans since IPMN and pancreatic cysts are \textit{hypo-intensity} regions in T1 scans. In contrast, we use maximum intensity projection for T2 scans, because IPMN and pancreatic cysts correspond to \textit{hyper-intensity} regions in these scans. The intensity projections corresponding to T1 and T2 scans can be represented as:

\begin{align}
\label{eqn:eqlabel}
\begin{split}
\mathcal{I'}=\minimum_{Z_1} \ [\mathcal{I}(X_1,Y_1,Z_1)],
\\
\mathcal{J'}=\maximum_{Z_2} \ [\mathcal{J}(X_2,Y_2,Z_2)],
\end{split}
\end{align}

 \noindent where $Z_1$ and $Z_2$ consists of $k$ slices around $I_u$ and $J_v$ respectively. Moreover, $\mathcal{I'}$ and $\mathcal{J'}$ represent the intensity projections from T1 and T2 scans, respectively. The overview of the proposed approach is shown in Figure~\ref{fig:workflow} \\

\noindent{\textbf{Network Architecture:}}\\
In order to obtain deep feature representation for our proposed IPMN classification approach, we use (fast) CNN-F architecture trained on ImageNet~\cite{chatfield2014return}. The architecture consists of 5 convolutional and 3 fully-connected layers. The input 2D image is resized to 224 $\times$ 224. The first convolutional layer contains 64 filters with stride 4 and there are 256 filters with stride 1 in the other 4 convolutional layers. Our input to the network are the 2D intensity projections $\mathcal{I'}$ and $\mathcal{J'}$, whereas the features are extracted from the second fully connected layer without applying non-linearities such as ReLU (Rectified Linear Units). The features are $\ell_2$ normalized to obtain the final representation.

\subsection {Feature Fusion with Canonical Correlation Analysis:}
The next step is to combine information from the two imaging modalities so as to improve the classification performance. As these two imaging modalities (T1 and T2) have complementary information, the fusion of features from these modalities can help improve IPMN diagnosis. Assume that $\Phi \in\mathbb{R}^{n \times p}$ and $\Psi \in\mathbb{R}^{n \times q}$ comprise the deep features from the intensity projections of $n$ training images from T1 and T2 scans respectively. Each sample has a corresponding binary label given by $\mathcal{Y}=[y_1,y_2 \dots y_n]$, where $\mathcal{Y}\in \{0,1\}^{n \times 1}$.  Consider $C_{\Phi \Phi} \in\mathbb{R}^{p \times p}$ and $C_{\Psi \Psi} \in\mathbb{R}^{q \times q}$ represent the within sets covariance matrices of $\Phi$ and $\Psi$ respectively. Additionally, the between set covariance matrix is referred as $C_{\Phi \Psi} \in\mathbb{R}^{p \times q}$ such that $C_{\Psi \Phi}=C_{\Phi \Psi}^T$. The covariance matrix $\mathcal{C}$ can therefore be written as:

\begin{equation}
\mathcal{C}=
\begin{pmatrix}
cov(\Phi) & cov(\Phi,\Psi)\\
cov(\Psi,\Phi) & cov(\Psi)\\
\end{pmatrix}=
\begin{pmatrix}
C_{\Phi \Phi} & C_{\Phi \Psi}\\
C_{\Psi \Phi} & C_{\Psi \Psi}\\
\end{pmatrix}
\end{equation}

In this regard, CCA is employed to find the linear combinations, $\Phi^*=W_{\Phi}^T\Phi$ and $\Psi^*=W_{\Psi}^T\Psi$ such that the pair-wise correlation between the two sets is maximized~\cite{haghighat2016fully}. CCA is a method that can help explore the relationship between the two multi-variate variables. The pairwise correlation between the two sets can be modeled as:

\begin{equation}
corr(\Phi^*,\Psi^*)=\frac{cov(\Phi^*,\Psi^*)}{var(\Phi^*).var(\Psi^*)},
\end{equation}

\noindent where $cov(\Phi^*,\Psi^*)=W_{\Phi}^TC_{\Phi \Psi}W_{\Psi}$, $var(\Phi^*)=W_{\Phi}^TC_{\Phi \Phi}W_{\Phi}$ and  $var(\Psi^*)=W_{\Psi}^TC_{\Psi \Psi}W_{\Psi}$. The covariances are then used to find the transformation matrices $W_{\Phi}$ and $W_{\Psi}$ using the following equations:

\begin{align}
\label{eqn:eign}
\begin{split}
C_{\Phi \Phi}^{-1}C_{\Phi \Psi}C_{\Psi \Psi}^{-1}C_{\Psi \Phi}\widetilde{W}_{\Phi}=\ \Lambda^2 \widetilde{W}_{\Phi},
\\
C_{\Psi \Psi}^{-1}C_{\Psi \Phi}C_{\Phi \Phi}^{-1}C_{\Phi \Psi}\widetilde{W}_{\Psi}=\ \Lambda^2 \widetilde{W}_{\Psi}.
\end{split}
\end{align}

\noindent In the above equation, $\widetilde{W}_{\Phi}$ and $\widetilde{W}_{\Psi}$ are the eigenvectors and $\Lambda^2$ is the eigenvalue diagonal matrix. 

Lastly, the final feature matrix can be represented as the sum of the transformed feature matrices from the two modalities:

\begin{equation}
\label{eq:final}
F={W}_{\Phi}^T\Phi+{W}_{\Psi}^T\Psi=\begin{pmatrix}
{W}_{\Phi}\\
{W}_{\Psi}
\end{pmatrix}^T\begin{pmatrix}
\Phi\\
\Psi
\end{pmatrix}.
\end{equation}

The learned transformation is also applied to the features from test images in order to obtain the final transformed testing features.

\section{Experiments and Results}
In order to account for the mis-alignment between T1-weighted and T2-weighted scans, we performed Multi-resolution image registration using image pyramids~\cite{ITKSoftwareGuideThirdEdition}. The registration results were examined and images with mis-registration were removed from the final evaluation set. Our final evaluation set comprised 139 scans from each modality and we performed 10 fold cross validation over the dataset. The minimum (maximum) intensity projection images from T1 (T2) scans were fed into the deep CNN-F network and feature representation from each of these images was used to obtain the final CCA based discriminative representation (Eq.~\ref{eq:final}). We then employed Support Vector Machine (SVM) classifier to obtain the final classification labels i.e. normal vs IPMN.

\begin{table}[t!]
\footnotesize
\caption{\textit{Results for accuracy, sensitivity and specificity of the proposed multi-modal fusion approach along with standard error of the mean (SEM) in comparison with single modality and feature concatenation based approaches.}}
\vspace{0.1 in}
\label{table:Results}
\begin{tabular}{l@{\hspace{0.1in}}c@{\hspace{0.15in}}c@{\hspace{0.15in}}c@{\hspace{0.05in}}c}
\toprule[1.5pt] \multirow{2}{*}{\textbf{Methods}}   & \multirow{2}{*}{\textbf{Accuracy}} & \multirow{2}{*}{\textbf{Sensitivity}} & \multirow{2}{*}{\textbf{Specificity}}\\ 
\multirow{2}{*}{}   & \multirow{2}{*}{(SEM \%)} & \multirow{2}{*}{(SEM \%)} & \multirow{2}{*}{(SEM \%)}\\
\\
\cmidrule(r){1-4}
T1-weighted    &      84.23 (1.10)  &	 	89.16 (0.88) 		& 55.00  (4.16)&        \\
T2-weighted &      61.04 (1.35) &		59.59 (2.25) &			 57.67 (2.95) &      \\
Concat. of T1 \& T2 &      82.09 (1.01) 	& 88.49 (0.90) 	& 49.33 (3.48) &        \\
\textbf{Feature Fusion (Proposed)}     &	\textbf{82.80 (1.17)} & \textbf{83.55 (1.13)} & \textbf{81.67 (2.53)} &       \\
\toprule[1.5pt]
\end{tabular}
\end{table}

We compared our proposed multi-modal feature fusion based approach with single modality and feature concatenation based approaches. Since there exists an imbalance between the number of positive and negative examples, we performed Adaptive Synthetic Sampling (ADASYN)~\cite{he2008adasyn} to generate synthetic samples. This sampling enabled to generate synthetic feature examples from the minority class (normal). 
\vspace{-0.2cm}
\begin{table}[h]
\begin{center}
\caption{\textit{Classification accuracy along with standard error of the mean (SEM) for three class classification (normal, low grade IPMN and high grade IPMN) of proposed approach in comparison with other approaches.
}}
\label{table:threeclass}
\vspace{0.1 in}
\begin{tabular}{l@{\hspace{0.3in}}c@{\hspace{0.1in}}c}
\toprule[1.5pt] \multirow{2}{*}{\textbf{Methods}}   & \multirow{2}{*}{\textbf{Accuracy \% (SEM \%)}} \\ 
\\
\cmidrule(r){1-3}
T1-weighted    &      58.30 (1.52)  &      \\
T2-weighted &      45.93 (1.60) &      \\
Concat. of T1 and T2 &   56.81 (1.51)           \\
\textbf{Feature Fusion (Proposed)} & \textbf{64.67 (0.83)}  &     \\
\toprule[1.5pt]
\end{tabular}
\end{center}\vspace{-0.1in}
\end{table}

Table~\ref{table:Results} shows the results of our proposed approach in varying conditions. It can be observed that the performance of our proposed approach significantly outperforms the single modality and feature concatenation based approaches. The T1 based classification yielded the highest sensitivity, but with very low \textit{specificity}. For IPMN classification, low specificity can be a serious problem as that can lead to unwarranted surgery and resection. In this regard, our proposed approach reports more than \textbf{30\% improvement} in specificity in comparison with the feature concatenation based approach.

\begin{figure}[t]
\centering
\includegraphics[width=80mm,height=95mm]{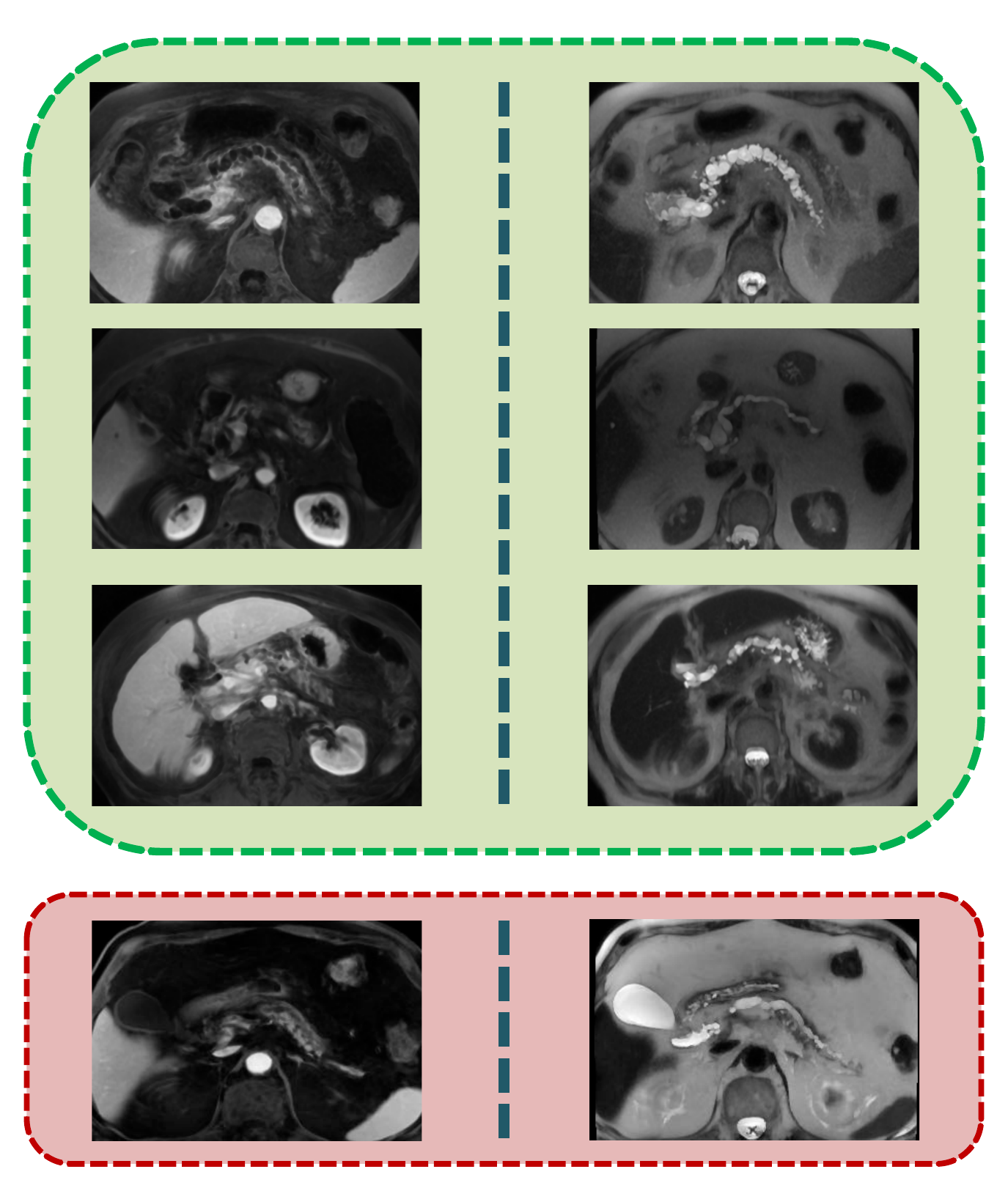}
\caption{Qualitative results of our proposed approach, showing minimum and maximum intensity projected images for T1 and T2 scans on left and right respectively. Each row represents a different case where the images correctly classified as IPMN are shown in green, whereas the misclassification of IPMN as normal are shown in red. \textit{(Edit: The misclassified cases now correctly correspond to T1 and T2 projections)}}
\label{fig:qual}
\vspace{-0.35 cm}
\end{figure}

It is important to note that since our proposed approach is based on the correlation and covariance in the data, it doesn't require explicit sample balancing using ADASYN. Moreover, for experiments, we also tried features from various layers of CNN-F as well as features from deeper residual networks such as ResNet-50, ResNet-101, and ResNet-152~\cite{he2016deep}. The best classification results, however, were obtained using the second fully-connected layer of CNN-F. Figure~\ref{fig:qual} shows the qualitative results of our proposed approach with intensity projections from T1 and T2 scans. The cases shown in green are the correctly classified as IPMN whereas those shown in red are incorrectly classified as normal. \\

\noindent\textbf{Low and High grade IPMN classification:}\\
We also performed 3-class classification using our proposed approach. Out of 108 IPMN subjects, 48 had low-grade IPMN whereas the remaining 60 had high-grade IPMN or invasive carcinoma. Using the features obtained from the CCA based fusion, we train a 3-class SVM classifier with classes normal, low-grade IPMN and high-grade IPMN. These diagnostic labels were obtained using the pathology report after surgery. Table \ref{table:threeclass} shows the performance of our proposed approach for normal, low-grade and high-grade IPMN classification. The proposed CCA based classification approach outperforms single modality and feature concatenation based approaches. The CCA based approach reports around \textbf{8\% improvement} in comparison to the feature concatenation based approach.

\section{Discussion and Conclusion}
Pancreatic cancer is projected to become the second leading cause of cancer-related deaths before 2030~\cite{rahib2014projecting}. IPMNs are the radiographically identifiable precursor to pancreatic cancer. In this paper, we proposed a multi-modal feature fusion framework to perform the classification of IPMN. Inspired by the clinical need to identify subjects with IPMN, our proposed approach can help radiologists in diagnosing invasive pancreatic carcinoma. In contrast to previous studies, this is the first approach to use deep CNN feature representation for IPMN diagnosis. Moreover, we empirically show the importance of feature level fusion of two different MRI imaging modalities i.e. T1 and T2 scans.

Another advantage of our proposed approach is that it doesn't require manual segmentation of pancreas or cysts as in other approaches. We only need to identify a single slice where pancreatic tissues can be prominently observed. Additionally, by using the intensity projections across a consecutive set of slices, we can develop robustness to the manual selection of a single slice. As the CCA is used to learn the transformation, its use also circumvents the need to have explicit data balancing in the cases of imbalance between positive and negative examples.

As an extension to this study, our future work will involve joint detection and diagnosis of IPMN in MRI scans. As the number of subjects undergoing screening for IPMN increases, we can get sufficient data to perform an end-to-end training or fine-tuning of a 3D convolutional neural network. The use of Generative Adversarial Networks (GANs)~\cite{goodfellow2014generative} can assist in data augmentation by generating realistic examples to further improve the training of the network.


Furthermore, the segmentation of pancreas and IPMN cysts can help in localizing the regions of interest. These regions can be used not only to extract discriminative imaging features, but also to extract important measurements such as the diameters of main pancreatic duct and cysts~\cite{tanaka2012international}. The inclusion of additional imaging modalities such as CT scans along with demographic and clinical characteristics, including age, gender, family history, symptoms and body fat can help further improve diagnostic decision making in the future.




\bibliographystyle{IEEEbib}
\bibliography{refs}

\end{document}